\documentclass[runningheads]{llncs}
\usepackage{xcolor}
\usepackage{microtype}
\usepackage{booktabs} 
\usepackage[loadonly]{enumitem} 
\usepackage{amsmath}
\usepackage{gensymb}
\usepackage{mathtools} 
\usepackage{extarrows} 
\usepackage{subfig}
\usepackage{nth}
\usepackage{microtype}
\usepackage{hyperref}
\usepackage{amssymb}
\usepackage{lineno}
\usepackage{algorithm}
\usepackage[noend]{algpseudocode}
\usepackage{listings}
\usepackage{paralist}
\usepackage{adjustbox}
\usepackage{soul}

\begin{document}

\title{Heterogeneous Relational Reasoning in Knowledge Graphs with Reinforcement Learning}

\author{Mandana Saebi\inst{1} \and
Steven Krieg\inst{1}\and
Chuxu Zhang\inst{1}\and Meng Jiang\inst{1} \and Nitesh Chawla\inst{1}}
\authorrunning{Saebi et al.}
\titlerunning{Heterogeneous Relational Reasoning in Knowledge Graphs}

\institute{University of Notre Dame, Notre Dame, IN, USA \\
\email{msaebi@nd.edu}, \email{skrieg@nd.edu}\\
\email{czhang@nd.edu}, \email{mjiang2@nd.edu}, \email{nchawla@nd.edu}\\
}

\maketitle

\newcommand{\ot}{Ours (-T)}
\newcommand{\on}{Ours (-N)}
\newcommand{\otn}{Ours (TN)}

\begin{abstract}
\textbf{Path-based relational reasoning over knowledge graphs has become increasingly popular due to a variety of downstream applications such as question answering in dialogue systems, fact prediction, and recommender systems. In recent years, reinforcement learning (RL) has provided solutions that are more interpretable and explainable than other deep learning models. However, these solutions still face several challenges, including large action space for the RL agent and accurate representation of entity neighborhood structure.  We address these problems by introducing a type-enhanced RL agent that uses the local neighborhood information for efficient path-based reasoning over knowledge graphs. Our solution uses graph neural network (GNN) for encoding the neighborhood information and utilizes entity types to prune the action space. Experiments on real-world dataset show that our method outperforms state-of-the-art RL methods and discovers more novel paths during the training procedure.}
\end{abstract}

\makeatletter
\makeatother
\section{Introduction}
\ifx
\begin{figure}
    \centering
    \includegraphics[width=0.5\linewidth]{figs/kglarge.png}
    \caption{A subgraph of the NELL-995 knowledge graph. Nodes, or entities, are colored according to their type. Edges, or relations, also have types, which are generally related to the types of the entities joined by the relation.}
    \label{fig:kglarge}
\end{figure}
\fi
Relational reasoning has been long one of the most desirable goals of machine learning and artificial intelligence~\cite{koller2007introduction,muggleton1991inductive,kemp2006learning,xu2012infinite}. In the context of large-scale knowledge graphs (KG), relational reasoning addresses a number of important applications, such as question answering~\cite{xiong2017deeppath,das2017go}, dialogue systems~\cite{serban2017deep,peng2017composite}, and recommender systems~\cite{zheng2018drn,liu2018deep,chen2019top}. Most KGs are incomplete, so the problem of inferring missing relations, or KG completion, has become an increasingly popular research domain. Several works view this as a link prediction problem and attempt to solve it using network embedding and deep learning approaches~\cite{bordes2013translating,socher2013reasoning,trouillon2016complex,jia2018knowledge,yang2014embedding,dettmers2018convolutional}. These methods embed the KG into a vector space and use a similarity measure to identify the entities that are likely to be connected. However, they are unable to discover the reasoning paths, which are important for interpreting the model. Furthermore, they do not provide an explicit explanation during the learning process and often rely on other analytical methods to provide an interpretation for their predictions. As a result, it is often hard to trust the predictions made by embedding-based methods.
Recent advances in the area of deep reinforcement learning (DRL) have inspired reinforcement learning (RL) based solutions for the KG completion problem~\cite{sharma2010graph,das2017go,xiong2017deeppath,lin2018multi,qu2018curriculum,shen2018m,lin2019rel4kc,xian2019reinforcement}. RL-based methods formulate the task of KG completion as a sequential decision-making process in which the goal is to train an RL agent to walk over the graph by taking a sequence of actions (i.e., choosing the next entity) that connects the source to the target entity. The sequences of entities and relations can be directly used as a logical reasoning path for interpreting model predictions. For example, in order to answer the query \textit{(Reggie Miller, plays sport, ?)}, the agent may find the following reasoning path in the KG: \textit{Reggie Miller} $\xrightarrow{\textit{competes with}}$ \textit{Michael Jordan} $\xrightarrow{\textit{plays sport}}$ \textit{Basketball}. In this case \textit{Reggie Miller}, \textit{Michael Jordan}, and \textit{Basketball} are all nodes (entities) in the KG, and \textit{competes with} and \textit{plays sport} are edges (relations). The agent is thus learning to navigate the nodes and edges of the KG.

These RL solutions demonstrate competitive accuracy with other deep learning methods while boasting improved interpretability of the reasoning process. However, there are some fundamental and open issues that we address in this work:
    \par {\em Large action space.} In KGs, facts are represented as binary relations between entities. Real-world KGs contain huge numbers of entities and relations. As a result, the RL agent often encounters nodes with a high out-degree, which increases the complexity of choosing the next action. In these cases, exploring the possible paths to determine the optimal action is computationally expensive, and in many cases beyond the memory limits of a single GPU. Previous studies have shown that type of information can improve the KG completion performance~\cite{shen2020modeling,lei2019path} using deep learning approaches. To improve the search efficiency, we first propose a representation for the entity type information, which we include in the representation of the state space. We then prune the action space based on the type information. This guides the RL agent to limit the search to the entities whose type best matches the previously taken actions and, as a result, avoid incorrect reasoning paths. In the above example reasoning path for query \textit{(Reggie Miller, plays sport, ?)}, suppose the entity \textit{Michael Jordan} has a high out-degree, and is connected to several other entities through different relations (e.g., \textit{Michael Jordan} $\xrightarrow{\textit{received award}}$ \textit{NBA Champion},\\ \textit{Michael Jordan} $\xrightarrow{\textit{born in}}$ \textit{New York City, Michael Jordan} $\xrightarrow{\textit{married to}}$ \textit{Juanita Vanoy}). None of these additional entities are useful for answering the query and may mislead the agent. However, we demonstrate that an agent can learn that next entity's type is most likely a \textit{sport} rather than a \textit{person} or a \textit{location}.

    \par {\em Accurate representation of entity neighborhood}. Existing RL-based methods for KG completion do not capture the entity's neighborhood information. Previous studies on one-shot fact prediction have shown that the local neighborhood structure improves the fact prediction performance for long-tailed relations~\cite{xiong2018one,zhang2019few}. We propose a graph neural network (GNN)~\cite{kipf2016semi} to encode the neighborhood information of the entities and leverage the state representation with the type and neighborhood information of the entities. We demonstrate that learning the local heterogeneous neighborhood information improves the performance of RL agent on the long-tailed relations, which in turn significantly improves performance for the KG completion task. 

Our contributions include:
\begin{enumerate}
    \item Designing an efficient vector representation of entity type-embeddings.
    \item Pruning the action space and improving the choice of next actions using the entity type information. 
    \item Proposing a GNN for incorporating the local neighborhood information in the state representation.
\end{enumerate}

The rest of the paper is organized as follows. First, in Section \ref{sec:related} we survey related work. Next, in Section \ref{sec:model} we present the details of our model, and in Section \ref{experiments} present and discuss experimental results. Finally, we conclude in Section \ref{sec:conclusion} and discuss opportunities for continued work.

\section{Related Work}
\label{sec:related}
Relational reasoning over knowledge graphs has attracted significant attention over the past few years. Recent works~\cite{bordes2013translating,socher2013reasoning,trouillon2016complex,jia2018knowledge,yang2014embedding,dettmers2018convolutional,yao2019graph} approached this problem by embedding the relation and entities into a vector space and identifying related entities by similarity in the vector space. However, these methods have some important drawbacks, including:
\begin{enumerate}
    \item They cannot perform multi-hop reasoning. That is, they only consider pairwise relationships and cannot reason along a path.
    \item They cannot explain the reasoning behind their predictions. Because they treat the task as a link prediction problem, the output of their prediction is binary.
\end{enumerate}

With the recent success of deep reinforcement learning in AlphaGO~\cite{silver2016mastering} researchers began to adopt RL to solve a variety of problems that were conventionally addressed by deep learning methods, such as ad recommendation~\cite{zheng2018drn,liu2018deep,chen2019top}, dialogue systems~\cite{serban2017deep,peng2017composite}, and question answering~\cite{xiong2017deeppath,das2017go}.  As a result, more recent methods proposed using RL to solve the multi-hop reasoning problem in knowledge graphs by framing it as a sequential decision-making process ~\cite{das2017go,shen2018reinforcewalk,xiong2017deeppath,shen2018m,li2018path,lin2018multi}. Deeppath~\cite{xiong2017deeppath} was the first method that used RL to find relation paths between two entities in KGs. It walks from the source entity, chooses a relation and translates to any entity in the tail entity set of the relation. MINERVA~\cite{das2017go}, on the other hand, learns to do multi-hop reasoning by jointly selecting an entity-relation pair via a policy network. MARLPaR ~\cite{li2018path} uses a multi-agent RL approach where two agents are used to perform the relation selection and entity selection iteratively. Lin et al. ~\cite{lin2018multi} implement reward shaping to address the problem of the sparse reward signal and action dropout to reduce the effect of incorrect paths. Xian et al.~\cite{xian2019reinforcement} use KG reasoning for recommender systems and designed both a multi-hop scoring function and a user-conditioned action pruning strategy to improve the efficiency of RL-based recommendation. 

Because these RL models treat the KG completion problem as a path reasoning problem instead of a link prediction problem, they are able to overcome both drawbacks of embedding methods that are outlined above. However, the RL models have drawbacks of their own, the most notable of which are computational cost and predictive accuracy. Many of these RL methods have tried to combine the representational power of embeddings and reasoning power of RL by training an agent to navigate an embedding space. For example, the authors of~\cite{lin2018multi} build an agent-based model on top of pre-trained embeddings generated by ComplEx\cite{trouillon2016complex} or ConvE~\cite{dettmers2018convolutional}. While we take a similar modular approach, our solution enriches the embedding space with additional information about entity types and local neighborhood information. 
In light of recent work on heterogeneous networks that have demonstrated the importance of heterogeneous information~\cite{dong2017metapath2vec,zhang2020learning,shen2020modeling,lei2019path} and local neighborhood information~\cite{xiong2018one,zhang2019few} in graph mining, we take a broader approach. We propose to include entity type information in the state representation to help the RL agent to take more informed actions considering the heterogeneous context. We also learn the heterogeneous neighborhood information simultaneously with training the RL agent to improve the prediction on less frequent relations.

\section{Model}
\label{sec:model}
\begin{figure*}
    \centering
    \includegraphics[width=\linewidth]{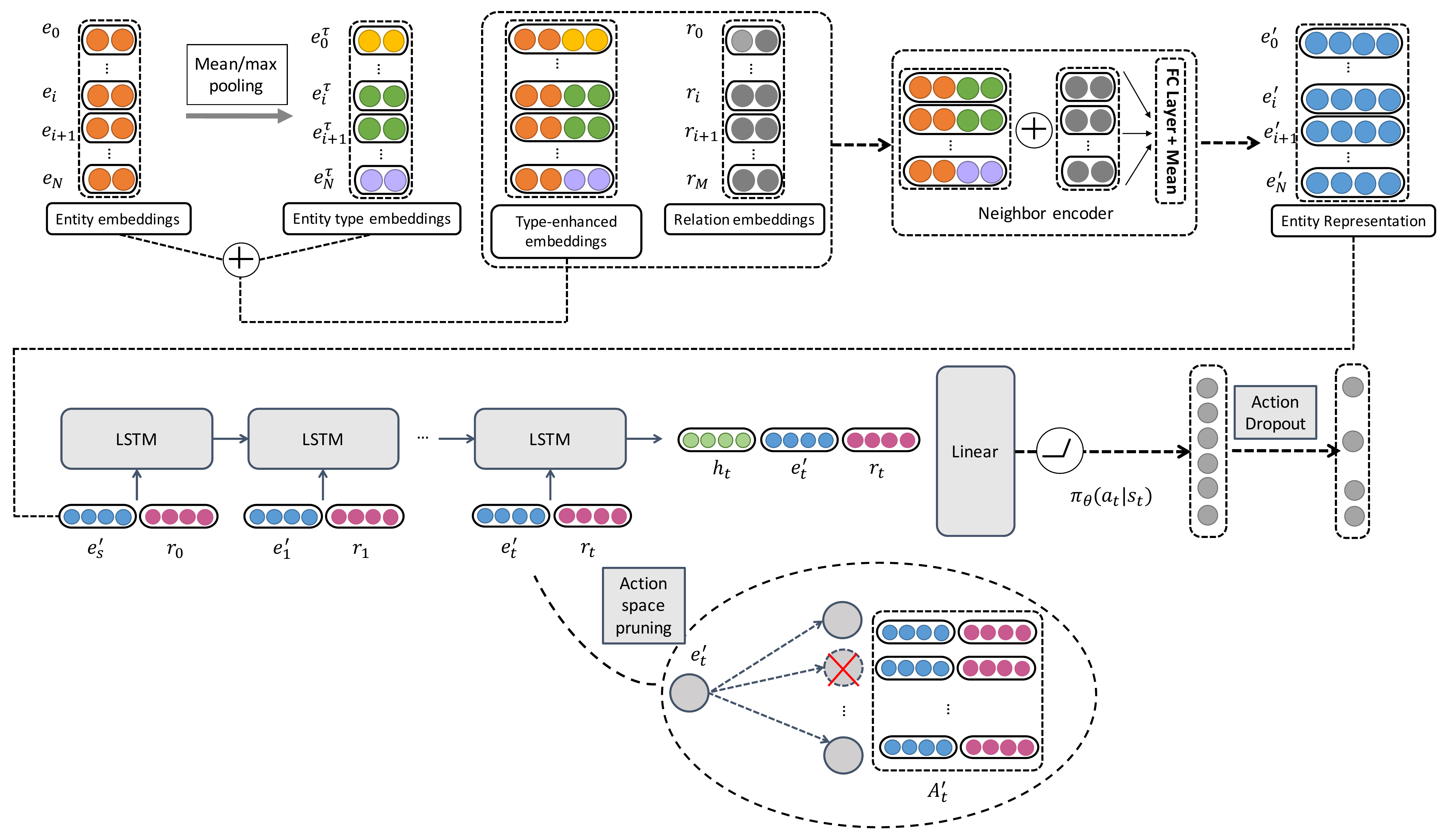}
    \caption{Model overview. The type embeddings are first created by max/mean pooling on the entities with a similar type. The type embeddings are then concatenated with the entity embeddings to create the type-enhanced embeddings, which are then passed to the neighbor encoder to create the final entity representation.}
    \label{fig:model}
\end{figure*}

In this section, we formally define the problem of relational reasoning in a KG and explain our RL solution. We then introduce our contribution by incorporating type-embeddings and heterogeneous neighbor-encoder. An overview of the model is displayed in Figure~\ref{fig:model}.

\subsection{Problem Formulation} \label{sec:problem}
Knowledge graphs consist of facts represented as triples. We formally define a knowledge graph $\mathcal{G} = \{(e_s,r,e_d)\} \subseteq \mathcal{E} \times \mathcal{R} \times \mathcal{E}$ where $\mathcal{E}$ is a set of entities and $\mathcal{R}$ is a set of relations. Given a query $(e_s,r,?)$, $e_s$ is called the source entity and $r$ is the query relation. Our goal is to predict the target entity $e_d \in \mathcal{E}$. In most cases, the output of each query is a list of candidate entities, $\hat{E_d} = \{\hat{e_1}, ..., \hat{e_n}\}$ for some fixed $n < |\mathcal{E}|$, ranked in descending order by probability. The prediction can be represented as a function $\mathcal{F}: \mathcal{E} \times \mathcal{R} \rightarrow \mathcal{E}^n$. 

In KGs, we are not only interested in accurate prediction of the target entity $e_d$, but also understanding the reasoning path the model uses to predict $e_d$. This is a key advantage that RL methods offer over embedding-based models, which can make entity predictions but cannot give interpretable justification for them. Rather than treating the task as a form of link prediction, RL models instead train an agent to traverse the nodes of a KG via logical reasoning paths. Below we provide the details of RL formulations of the problem.

\subsection{A Reinforcement Learning Solution} \label{sec:problem}
Similar to ~\cite{das2017go,lin2018multi,xiong2017deeppath}, we formulate this problem as a Markov Decision Process (MDP), in which the goal is to train a policy gradient agent (using REINFORCE~\cite{williams1992simple}) to learn an optimal reasoning path to answer a given query $(e_s,r,?)$. We express the RL framework as a set of states, actions, rewards, and transitions.

\textit{\textbf{States.}}
The state $s_t$ at time $t$ is represented as tuple $((e_s,r),e_t,h_t)$, where $(e_s, r)$ is the input query, $e_t$ is the entity at which the agent is located at time $t$ and $e_t$ is the history of the entities and relations traversed by agent until time $t$. The agent begins at the source entity with initial state $s_0 = ((e_s, r), e_s,h_0)$. We refer to the terminal state as $s_T = ((e_s, r), e_T,h_T)$, where $e_T$ is the agent's answer to the input query and $h_T$ is the full reasoning path. Each entity and relation is represented by an embedding vector $\mathbb{R}^d$ for some constant $d \in \mathbb{Z}$. In our solution, we enrich the state representation with entity type and neighborhood information, which is explain later in Section~\ref{entity_rep}.

\textit{\textbf{Actions.}}
At each time-step, the action is to select an edge (and move to the connecting entity) or stay at the current entity. The action space $A_t \subseteq A$ given state $s_t$ is thus set of all immediate neighbors of the current node $e_t$, and the node itself, i.e., $A_t(s_t) = \mathcal{N}_{e_t} \cup {e_t}$, where $\mathcal{N}_e$ is the set of all neighbors from node $e$. The inclusion of the current node $e_t$ in the action space represents the agent's decision to terminate, and so select $e_t$ as its answer to the input query. Since the graph is directed, $\mathcal{N}$ only includes nodes adjacent on out-edges. Following previous work~\cite{bordes2013translating,das2017go,xiong2017deeppath}, for each edge (triple) $(e_s,r,e_d)$, during preprocessing we add an inverse edge $(e_d,r^{-1},e_s)$ in order to facilitate graph traversal.

Most real-world knowledge graphs are scale-free, meaning that a small percentage number of entities have a high out-degree while the majority of the entities have a small out-degree. However, the entities with high out-degree are crucial to query answering. For performance reasons, many RL models are forced to cap the size of the action space and do so via a pre-computed heuristic. For example, \cite{lin2018multi} pre-computes PageRank scores for each node, and narrows the action space to a fixed number of highest-ranking neighbors. In this work, we use entity type information to limit the search to the entities with best-matching types, given the previous actions. We call the reduced action space $A'_t$. We provide more details in Section~\ref{type_embeddings}.

\textit{\textbf{Rewards.}}
The agent evaluates each action and chooses the one that will maximize a reward. Previous works~\cite{xiong2017deeppath,das2017go} define only a terminal reward of $+1$ if the agent reaches the correct answer. However, since knowledge graphs are incomplete, a binary reward cannot model the potentially missing facts. As a result, the agent receives low-quality rewards as it explores the environment. Inspired by ~\cite{lin2018multi}, we use pre-trained KG embeddings based on existing KG embedding methods to design a soft reward function for the terminal state $s_T$ based on~\cite{ng1999policy}:

\begin{equation}
 R_T(s_T) =
    \begin{cases}
    1 & \text{if $(e_s, r, e_T) \in \mathcal{G}$}\\
    f(e_s, r, e_T) & \text{otherwise}\\ 
    \end{cases} 
\end{equation}
Where $f(e_s, r, e_T)$ is a similarity measure calculated based on pre-trained KG embedding approach~\cite{lin2018multi}. We use different embedding methods depending on different datasets. More details is provided in Section~\ref{experiments}.

\textit{\textbf{Transitions.}}
At state $s_t$, the agent choses an action  $a_t \in A_t$ based on a policy $\pi : \mathcal{S} \rightarrow \mathcal{A}$ where $\mathcal{S}$ and $\mathcal{A}$ are the sets of all possible states and and actions, respectively. Following \cite{lin2018multi}, we use an $LSTM$ to encode the history $h_t=\{e_{t-k},r_{t-k+1},...,e_{t-1},r_{t}\}$ of the past $k$ steps taken by the agent in solving the query. The history embedding for $h_t$ is represented as:
\begin{equation}
    \mathbf{h_t} = LSTM(\mathbf{h_{t-1}},a_{t-1})
\end{equation}

We define the policy network $\pi$ with weight parameters $\theta$ as follows:
\begin{equation}
    \pi_{\theta}(a_t|s_t) = \text{softmax}(A'_t \times W_2 ReLU(W_1[e_t; h_t; r])),
\end{equation}

The transition to a new state is thus given by:
\begin{equation}
    s_{t+1} = ((e_s, r), \operatorname*{argmax}_{a'_t \in A'_t}\pi_{\theta}(a'_t|s_t))
\end{equation}

To reduce the potential impact of $argmax$ leading to the overuse of incorrect paths, we utilize random action dropout as described in~\cite{lin2018multi}.

The policy network is trained using stochastic gradient descent to maximize the expected reward: 
\begin{equation}
    J (\theta) = \mathbb{E}_{(e_s,r,e_d)\in G}\mathbb{E}_{a_1,\dots,a_T \sim\pi_{\theta}} [R_T(s_T)]
\end{equation}

\subsection{Entity Types}
\label{type_embeddings}

Many knowledge graphs contain rich semantic information as entity type that can be used as prior information to guide the agent through the reasoning process. We argue that the type information can be helpful for reducing the action space, especially for nodes with a high out-degree. Entity type information can be used to limit the search only to the entities that are best matching the previously visited entities and actions. To achieve this, we measure the similarity of all possible actions given the entity type embedding of current and possible target entities and only keep the top $n$ candidates.  In order to build the entity type representation $e^{\tau}$, we propose to aggregate the vector representation of the entities with a similar type. Below we propose two simple mechanisms for doing so:

\begin{enumerate}
    \item Take the average of the embedding vectors for all the $N$ entities $e_i$ that share the same entity type $\tau$ (mean-pooling).
    \item Take the maximum value of each element in the the embedding vectors for all the $N$ entities $e_i$ that share the same entity type (max-pooling).
\end{enumerate}

To measure the similarity of the current entity with the candidate actions we use the cosine similarity of two entities with respect to their type-enhanced embedding vectors:

\begin{equation}\label{eq:one-hop-score}
\begin{aligned}
& g(e_0,e_k) = \\ 
& \left\langle\mathbf{[e_0\oplus{e_0}^{\tau}]},\mathbf{r_1},~\mathbf{[e_k\oplus{e_k}^{\tau}]}\right\rangle + b_{[e_k;{e_k}^{\tau}]}
\end{aligned}
\end{equation}
where $\langle\cdot,\cdot\rangle$ is the dot product operation, $\oplus$ represents the vector concatenation operator, $\mathbf{e}$, $\mathbf{r}\in\mathbf{R}^d$ are $d$-dimensional vector representations of the entity $e$ and relation $r$, and $b_{e}\in\mathbf{R}$ is a bias term for $e_k$. We call ${e_i}^{\tau'}=[e_i;{e_i}^{\tau}]$ the type-enhanced entity representation. We then rank the possible actions and prioritizes the ones that are more likely to result in a correct answer based on the $g(e_0,e_k)$ score. We thus create a pruned action space $A'_t$ by keeping the nodes with the highest value of $g$.

\subsection{Heterogeneous Neighbor Encoder}
\label{neighbor_encoder}
After generating the type embeddings, we feed the type-enhanced embeddings together with the relation embeddings to the heterogeneous neighbor encoder to generate the enriched entity representation. Although many works~\cite{bordes2013translating,yang2015embedding} have been proposed to learn entity embeddings using relational information, recent studies~\cite{xiong2018one,zhang2019few} have demonstrated that explicitly encoding graph local structure can benefit entity embedding learning in knowledge graph. Inspired by this motivation, we propose a heterogeneous neighbor encoder to learn the enriched entity embedding by aggregating entity's neighbors information. Specifically, we denote the set of relational neighbors ($relation, entity$) of given entity $h$ as $\mathcal{N}_{h} = \left \{ (r_{i}, t_{i}) | (h, r_{i}, t_{i})\in \mathcal{G}\right \}$, where $\mathcal{G}$ is the background knowledge graph, $r_{i}$ and $t_{i}$ represent the $i$-th relation and corresponding neighboring entity of $h$, respectively. The heterogeneous neighbor encoder should be able to encode $\mathcal{N}_{h}$ and output a feature representation of $h$ by considering different relational neighbors $(r_{i}, t_{i}) \in \mathcal{N}_{h}$. To achieve this goal, we formulate the enriched entity embedding as follows:
\begin{equation}\label{eq:neighbor_encoder}
\begin{aligned}
& f(h) = \sigma \left \{ \frac{1}{\mathcal{N}_{e}}\sum _{i} \left. \big(\mathcal{W}_{rt}(e_{r_{i}}^{\tau'}\oplus e_{t_{i}}^{\tau'}) + b_{rt}\right. \big) \right \}\\ 
\end{aligned}
\end{equation}
where $\sigma$ denotes activation unit (we use Tanh), $\oplus$ represents concatenation operator, $e_{t_{i}}^{\tau'}$, $e_{r_{i}}^{\tau'} \in \mathbb{R}^{d \times 1}$ are type-enhanced entity embeddings of $t_{i}$ and $r_{i}$. Besides, $u_{rt} \in \mathbb{R}^{d \times 1}$, $\mathcal{W}_{rt} \in \mathbb{R}^{d \times 2d}$ and $b_{rt} \in \mathbb{R}^{d \times 1}$ ($d$: pre-trained embedding dimension) are  parameters of neighbor encoder.


\section{Experiments}
\label{experiments}
In this section, we describe and discuss the experimental results of our proposed approach. We compare against several baseline methods: ConvE (embedding-based) \cite{dettmers2018convolutional}, ComplEx (embedding-based) \cite{trouillon2016complex}, MINERVA (agent-based) \cite{das2017go}, and MultiHopKG (agent-based) using both ConvE and ComplEx for pre-trained embeddings \cite{lin2018multi}.

\label{entity_rep}
\subsection{Data \& Metrics} \label{sec:data}
\vspace{-10pt}
\begin{table*}
    \small
    \begin{adjustbox}{max width=1\textwidth,center}
    \centering
    \setlength\tabcolsep{4pt}
    \begin{tabular}{|l|r|r|r|r|r|r|r|}
        \hline
        \textbf{Dataset} & \textbf{|$\mathcal{E}$|}  & \textbf{\#Types} & \textbf{|$\mathcal{R}$|} & \textbf{\#Facts} & \textbf{\#Queries} & \textbf{\#Queries Discarded} \\ \hline 
        NELL-995 & 75,492 & 268 & 200 & 154,213 & 3,992 & 1,152 \\ \hline 
        Amazon Beauty & 16,345 & 5 & 7 & 52,516 & 1,325 & 174 \\ \hline   
        Amazon Cellphones & 13,837 & 5 & 7 & 31,034 & 951 & 205 \\ \hline   
        \end{tabular}
    \end{adjustbox}
    \caption{Description of the data sets used for our experiments. $|\mathcal{E}|$ describes the total number of nodes in the KG, Types describes the number of entity types, $|\mathcal{R}|$ describes the number of edge types, and Facts describes the total number of edges. Queries is the test set, a subset of Facts which are removed from the KG for testing. Queries Discarded is the subset of Queries for which at least one of the entities does not appear in the training set.}
    \label{tab:data}
\end{table*}
\vspace{-20pt}

The experiments utilize the 3 data sets presented in Table~\ref{tab:data}.
Of the standard data sets used in KG reasoning tasks, NELL-995 is the only one that explicitly encodes entity types. Therefore, in addition to NELL, we incorporated two datasets from the Amazon e-commerce collection~\cite{he2016ups}. Each Amazon data contains a set of users, products, brands, and other information, which the authors of~\cite{xian2019reinforcement} use to make product recommendations to users. This task is a specialized instance of KG completion that only focuses on user-product relations, so we do not include it in our baseline results. Additionally, we found these data sets were too large for efficient computation in the broader KG completion task, so we shrunk them for our experiments. To do this, we randomly chose 20\% of the nodes and induced a subgraph on those nodes. While this might result in a sparser graph that makes predictions more difficult, this was the best option given the lack of other relevant data containing type information.

The full KG is represented by the number of $Facts$ in Table~\ref{tab:data}. Before training, we partition $Facts$ into a training set and a test set, which we call $Queries$. In NELL-995, this split already exists as part of the standard data set. For the Amazon sets, we populated $Queries$ with 10\% of the triples in $Facts$, chosen at random. Each model is then trained on the set $(Facts - Queries)$, and tested on the set $Queries$. Recall that each fact is a triple in the form $(e_s, r, e_d)$. Each triple is presented to the model in the form $(e_s, r, ?)$, and, as described in Section \ref{sec:problem}, the model outputs a list of ranked candidate entities $\hat{E_d} = \{e_1, ..., e_n\}$. Also recall that we describe the prediction as a function $\mathcal{F}: (e_s, r) \rightarrow \hat{E_d}$.

We measure performance for each experiment with standard KG completion metrics, namely, Hits@k for k=\{1,3,5,10\}, and Mean Reciprocal Rank (MRR). Hits@k is measured as the percentage of test cases in which the correct entity $e_d$ appears in the top $k$ candidates in $\hat{E_d}$, i.e.:
\begin{equation}
    Hits@k = \frac{|\{(e_s, r, e_d) \in Q : rank(e_d, \mathcal{F}(e_s, r)) \leq k\}|}{|Q|} \times 100
\end{equation}

where $Q = Queries$ and $rank(e_d, \hat{E_d})$ is a function that returns the position of entity $e_d$ in the set of ordered predictions $\hat{E_d}$. MRR is a related metric, defined as the multiplicative inverse of the rank of the correct answer, i.e.:
\begin{equation}
    MRR = \frac{1}{|Q|}\sum_{(e_s, r, e_d) \in Q}{\frac{1}{e_d, \mathcal{F}(e_s, r)}} \times 100
\end{equation}

 Because none of these models generalize to unknown entities, followed by previous work~\cite{das2017go,lin2018multi}, we measure Hits@k and MRR only for queries for which both $e_s$ and $e_d$ have already been seen at least once by the model during training. In other words, if either of the query entities is missing from the training set $(Facts - Queries)$, we discard it from testing. Additionally, we reserve a small portion of the $Facts$ as a development set to estimate performance during training.

\subsection{Parameter Selection} \label{sec:params}
For NELL-995, utilize the same hyperparameters described in \cite{lin2018multi} when training ConvE, ComplEx, Distmult, and Lin et al~\cite{lin2018multi} baselines. For MINERVA, we utilize the same hyperparameters described in \cite{das2017go} and train the model for 3,000 epochs. For the two Amazon datasets, we perform a grid search for our method and all baselines and report the best performance for each. For all datasets, we train the KG embedding models (ConvE and ComplEx) for 1000 epochs each. These embeddings are then used to make predictions directly but also serve as pre-trained inputs for the RL agent, which we train for 30 epochs per experiment for all datasets. We tried different embedding methods for the pre-trained embeddings and eventually used ComplEx for NELL-995 and Distmult for Amazon cellphones and Amazon Beauty as they resulted in the best performance in each data.

\begin{table*}
 \small 
    \begin{adjustbox}{max width=1.5\textwidth,center}
    \centering
    \begin{tabular}{|l|rrrr|rrrr|rrrr|}
        \hline
        \textbf{Data Set} & \multicolumn{4}{c|}{\textbf{NELL-995}} & \multicolumn{4}{c|}{\textbf{Amazon Beauty}} & \multicolumn{4}{c|}{\textbf{Amazon Cellphones}} \\
        
        Metric (\%) & 
        @1  & @5 & @10 & MRR & 
        @1  & @5 & @10 & MRR & 
        @1  & @5 & @10 & MRR \\ \hline
        
        ConvE & 
        \underline{68.2}   & \underline{85.4} & \underline{88.6} & \underline{76.1} & 
        25.8            & 44.9    & 55.3          & 35.2          &  
        16.9            & 35.0    & 44.8          & 25.7             \\  
        
        ComplEx 
        & 63.0         & 82.2 & 86.0 & 71.8 &
        \underline{27.6}  & 48.0 & 58.0 & \underline{37.5} &
        17.4           & 36.1 & 45.2 & 26.6 \\
        
        DistMult 
        & 65.1          & 83.5              & 85.7          & 73.4 &
        26.7            & \underline{48.8}     & \underline{58.1} & 37.1 & 
        \underline{18.8}   & \underline{39.7}     & \underline{48.4} & \underline{28.8} \\ \hline
        
        Lin et. al (2018) & 
        65.6    & 80.4  & 84.4 & 72.7 &
        20.6    & 33.6  & 39.5 & 27.1 & 
        12.2     & 22.2  & 27.6 & 17.5 \\
 
        MINERVA & 
        59.8    & 79.5 & 82.1 & 68.9 & 
        17.5    & 29.3 & 38.2 & 24.3 &  
        6.8     & 12.3 & 22.7 & 11.6 \\  
    
        \ot &
        66.9    & 81.9 &    85.2    & 74.1 &
        21.2    & 34.6 & 40.5 & 27.9 &
        12.6    &    22.9 &    28.1 &    17.9\\
        
        \on & 
        67.1    & 81.8      &85.1    &73.1 &
        20.7    & 33.8      &39.6    &27.2 & 
        12.3    &22.4       &27.9    &17.6\\
        
        \otn & 
        \textbf{68.9}   &\textbf{83.2}    & \textbf{86.7}    &\textbf{74.8} &
        \textbf{21.8}   & \textbf{35.1} & \textbf{40.7} & \textbf{28.2} & 
        \textbf{12.9}   & \textbf{23.1} & \textbf{28.5} & \textbf{18.2} \\
        \hline
    \end{tabular}
    \end{adjustbox}
    \caption{Experimental results on the NELL-995, Amazon Beauty, and Amazon Cellphones data sets. @\{1, 3, 5, 10\} and MRR are standard KG completion metrics and are described in Section \ref{sec:data}. The methods are separated into embedding-based (ConvE, ComplEx, and DistMult) and agent-based (MINERVA, Lin et al., and \otn) groups. Bolded numbers indicate the best-performing method from each group, with respect to the given metric.}
    \label{tab:results}
\end{table*}

\subsection{Experimental Results} \label{sec:results}
Our experimental results are described in Table \ref{tab:results}. For NELL-995 data, We quote the results reported in \cite{das2017go,lin2018multi}. Embedding-based methods show an overall higher performance compared to the RL-based methods. We can see that in all three datasets, our results outperform both RL baselines (Lin et al.~\cite{lin2018multi} and MINERVA~\cite{das2017go}). Amazon datasets, on the other hand, are far more challenging. We notice that even the embedding based methods are struggling with low performance on these datasets. On the Amazon data, the performance of all methods is significantly lower. Our method results in a 4\% improvement in MRR (and  5.43\% in Hits@1) over the best RL baseline on Amazon Cellphones and a 3.9\% improvement in MRR (and 5.5\% in Hits@1) on  Amazon Beauty. On the NELL-995 dataset, our method results in 2.8\% improvement in MRR and 4\% improvement in Hits@1 over the best performing baseline. We also performed ablations studies to analyze the effect of each module in our model. We removed the type embeddings in \ot~and the heterogeneous neighbor encoder in \on. We notice that removing the heterogeneous neighbor encoder results in a higher drop in performance in the Amazon datasets. This gap is quite smaller in the NELL-955 data.

Our results show that pruning the action space based on the entity type information results in a larger boost in performance on the Amazon datasets. We believe due to the sparsity of these two knowledge graphs, type information was more effective for action space pruning than entity page ranks, as done in~\cite{lin2018multi}. Note that, there are only 5 entity types in the Amazon datasets. As a result, the number of entities that will be discarded (due to type mismatch) is higher which assists the agent to discover a better path. We generated the type embeddings using max-pooling for the NELL-995 dataset and mean-pooling for both Amazon datasets. 

\subsection{Path diversity and convergence}
\begin{figure*}
    \centering
    \includegraphics[width=0.97\linewidth]{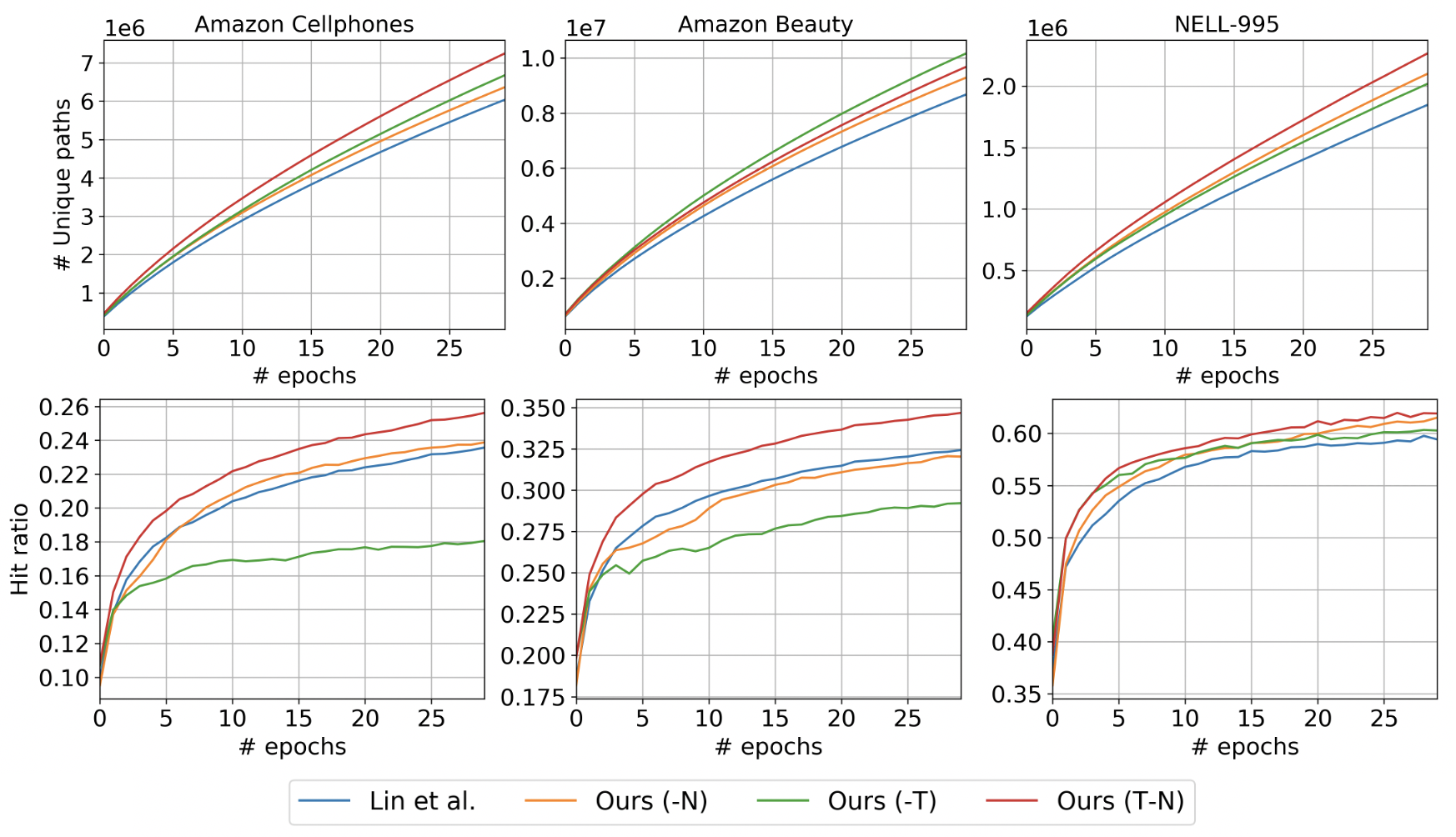}
    \caption{Performance of our model on the development set at different training epochs. The top figures show the number of unique paths visited during each epoch. The bottom rows show the hit ratio for the entire development set. }
    \label{fig:paths}
\end{figure*}

We compare the number of unique paths discovered from the development set during the training procedure. Figure~\ref{fig:paths} shows that path diversity (top row) improves across all models as the model performance (bottom row) improves. For this analysis, we compare our ablation models (\on~and \ot) with the best performing RL baseline by Lin et al.~\cite{lin2018multi}. Our method is more successful in discovering novel paths and obtains a better hit ratio on the development set. On the Amazon Beauty data, the number of unique paths discovered by \ot ~is higher than both combined (Ours) while in Amazon Cellphones the combined model performs better, but similar to Amazon Beauty, \on performs better than \ot. NELL-955 shows a different trend where removing the type information results in a larger drop in the number of unique paths, compared to the heterogeneous neighbor encoder. This is intuitive, since NELL-955 contains far more entity type than Amazon datasets, and inclusion of type information may be a positive factor for discovering new paths. In terms of convergence, Amazon Beauty and Amazon Cellphones show a similar trend, and removing the type information significantly reduces the hit ratio. This gap is smaller for NELL-995 data, though our model still shows improvement in hit ratio on this dataset.
 
 \vspace{-17pt}

\begin{table*}
\small 
    \begin{adjustbox}{max width=1.5\textwidth,center}
    \centering
    \begin{tabular}{|l|r|rrrr|r|rrrr|}
        \hline
        \textbf{Data Set} & \multicolumn{5}{c|}{\textbf{Seen Queries}} & \multicolumn{5}{c|}{\textbf{Unseen Queries}}\\
        \hline
        & 
        \% & \otn & \ot & \on &  Lin et al. &
        \% & \otn & \ot & \on &  Lin et al. \\ \hline

        NELL-995        & 15.3 & \textbf{53.3} & 53.2 (0)   & 51.1 (-5)     & 51.4 (-4) & 84.7 & \textbf{88.6} &87.4 (-1)          & 86.2 (-3)     & 85.5 (-3) \\
        Amazon-Beauty   & 89.6 & \textbf{25.5} & 23.8 (-7)  &  21.7 (-15)   & 23.9 (-6) & 10.4 & 39.1          & \textbf{39.2(0)}  &33.5  (-14)    & 36.6 (-7) \\
        Amazon-Cellphone& 87.9 & \textbf{20.2} & 18.1 (-10) & 15.6 (-23)    & 20.2 (-9) & 12.1 & \textbf{25.3} & \textbf{25.3 (0)} & 23.1 (-8)     & 22.8 (-10) \\
         \hline
    \end{tabular}
    \end{adjustbox}
    \caption{MRR on three datasets for queries of the development set that are seen / unseen in the training set. The percentage of seen / unseen queries in the development set for each data is shown in column \%.}
    \label{tab:seen-unseen}
\end{table*}

\begin{table*}
\small 
    \begin{adjustbox}{max width=1.5\textwidth,center}
    \centering
    \begin{tabular}{|l|r|rrrr|r|rrrr|}
        \hline
        \textbf{Data Set} & \multicolumn{5}{c|}{\textbf{To-Many}} & \multicolumn{5}{c|}{\textbf{To-One}}\\
        \hline
        & 
        \% & \otn & \ot & \on &  Lin et al. &
        \% & \otn & \ot & \on &  Lin et al. \\ \hline
        NELL-995            & 12.9 & \textbf{57.4} & 56.9 (0)   & 55.2 (0)      & 55.7 (0)      & 87.1  & \textbf{82.8}     &  82.0 (0)         & 81.0 (-1)     & 81.4 (-1) \\
        Amazon-Beauty       & 89.6 & \textbf{25.5} & 23.8 (-7)  & 21.7 (-15)    & 24.2 (-5)     & 10.4  & \textbf{39.1}     & 39.2 (0)          & 33.5  (-14)   & 36.8 (-6) \\
        Amazon-Cellphone    & 95.5 & \textbf{17.8} & 15.9 (-11) & 13.6 (-24)    & 16.5 (-7)    & 4.5   & \textbf{83.4}     & \textbf{83.4 (0)} & 66.6 (-20)    & 78.2 (-6)\\
         \hline
    \end{tabular}
    \end{adjustbox}
    \caption{MRR on three datasets for different relation types. The percentage of one-to-one and one-to-many relations in the development set for each data is shown in the \% column.}
    \label{tab:relation_types}
\end{table*}

\subsection{Performance on seen and unseen queries }
We compare the ablation models along with the best RL baseline performance on seen and unseen queries. Note that, percentage of unseen queries is much lower in the Amazon datasets compared to the NELL-995 dataset. Table~\ref{tab:relation_types} shows that our proposed method performs better on both seen and unseen queries. In particular, we notice that removing the neighbor encoder in Amazon Beauty and Amazon Cellphone results in a significant performance drop on unseen queries, while removing the type information had little or no effect. We observe a similar trend on the seen queries. Although they show a performance drop after removing the type information, the effect is less than removing the neighbor encoder.

\subsection{Performance on different relations}
We evaluate our proposed model on different relation types and compare our results with the best performing RL baseline. We take a similar approach as~\cite{lin2018multi} to extract \textit{to-many} and \textit{to-one} relations. A relation $r$ is considered \textit{to-many} if queries containing relation $r$ can have more than 1 correct answer, otherwise, it is considered a \textit{to-one} relation. Table~\ref{tab:relation_types} shows the MRR values on the development set for all three datasets. We notice that most relations in the Amazon datasets are \textit{to-many}, as these graphs are denser. On the other hand, a large portion of the NELL-995 data consists of \textit{to-one} relations. Overall, \textit{to-many} relations show lower performance, regardless of the model. Our proposed model consistently shows a better performance than Lin et al., except for the NELL-995 dataset where the improvement is marginal. Both \textit{to-one} and \textit{to-many} are more sensitive to removing the neighbor-encoder rather than removing the type information. However, for \textit{to-one} relations MRR does not drop in any dataset after removing the type information. 


\subsection{Case study}
\begin{figure}
    \centering
    \includegraphics[width=\linewidth]{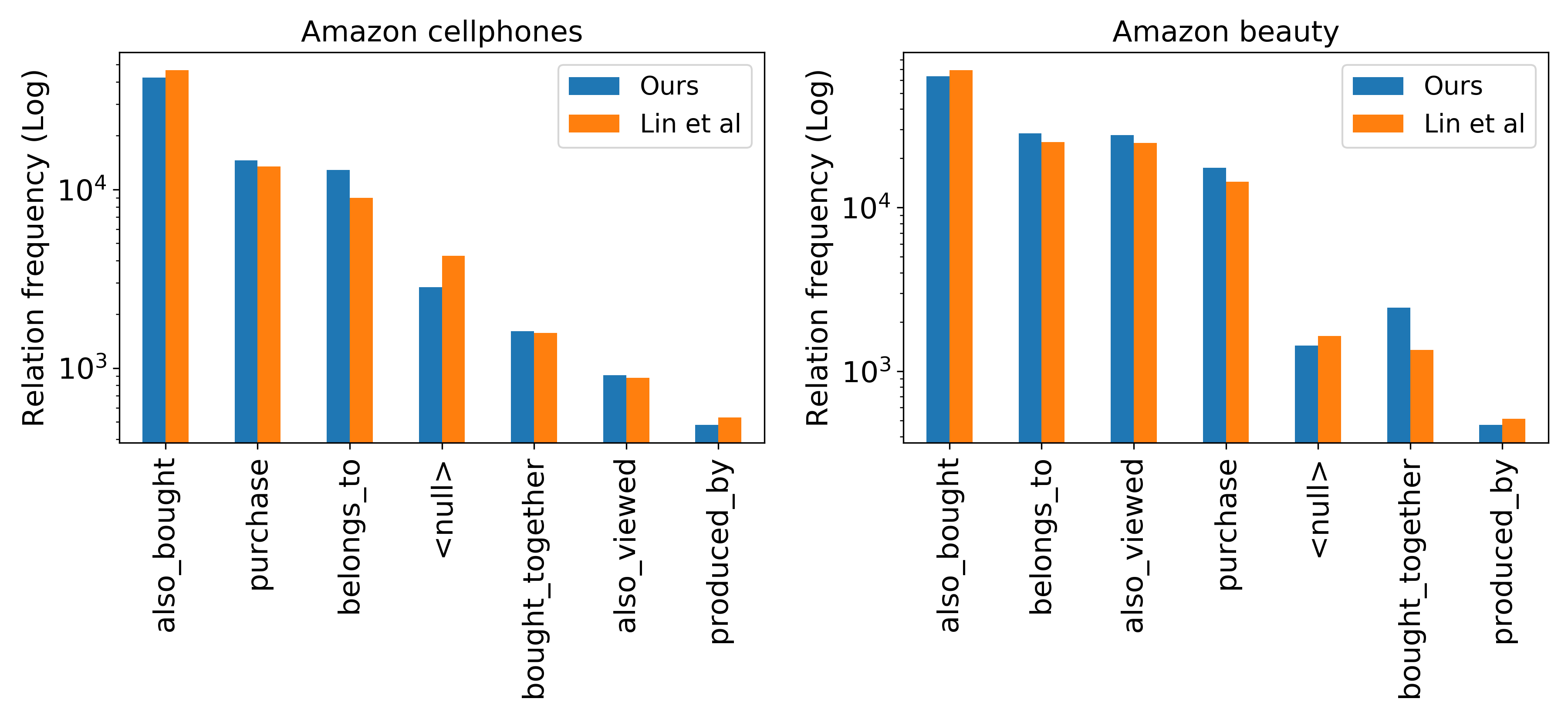}
    \caption{Relation frequencies in the discovered paths of the development set for the Amazon dataset. Note the log scale for the y-axis.}
    \label{fig:rel_freq_dev}
\end{figure}

In this section we present a few case studies that show the strength of our proposed method. In the NELL-995 dataset, our method is more successful when the agent encounters an entity with a high out-degree. As an example, for the query (\textit{Buffalo Bills (sports team)} , \textit{organization hired person} , ?) our method discovers the path: \textit{Buffalo Bills (sports team)} [D:26] $\xrightarrow{\text{coaches team(-1)}}$ \textit{Mike Mularkey (coach)} [D:5] $\xrightarrow{\text{coaches in league}}$ \textit{NFL (sports league)} [D:315] $\xrightarrow{\text{coaches in league(-1)}}$ \textit{Dick Jauron (coach)} [D:6], in which D is equivalent to the entity out-degree. While other baselines also find the partial path \textit{Buffalo Bills (sports team)} [D:26] $\xrightarrow{\text{coaches team(-1)}}$ \textit{Mike Mularkey (Coach)} [D:5] $\xrightarrow{\text{coaches in league}}$ \textit{NFL (sports team)} [D:315], they are not able to navigate properly after reaching the \textit{NFL} entity, due to its high out-degree. As a result, they are unable to discover the answer \textit{Mike Mularkey}. As another example,  we consider the query: (\textit{New York (City)}, $\textit{organization hired person}$ ,?). Our method discovers the path:  \textit{New York (City)} [D:314] $\xrightarrow{\text{subpart of(-1)}}$ \textit{Lincoln Center (attraction)} [D:5] $\xrightarrow{\text{location located within location}}$ \textit{NYC metropolitan area (island)} [D:80] $\xrightarrow{\text{at location(-1)}}$ \textit{Michael Bloomberg (politician)} [D:4]. Again, other RL baselines struggle with finding the next best step after entity \textit{New York}. Our method uses the location information to find the answer \textit{Michael Bloomberg}.
In the Amazon datasets, there are fewer entity and relation types. As a result, we observe many frequent patterns that all RL baselines are able to discover. Therefore, we focus on the diversity of the relations used in our method and the best performing baseline~\cite{lin2018multi} for the discovered paths in the development set. Figure~\ref{fig:rel_freq_dev} displays the inference results. On the Amazon cellphones data, our method discovers fewer {\em $<$null$>$}, {\em produced-by} and {\em also-bought} relations while it utilizes  more of other relations, in particular, {\em belongs-to} relations. Similarly, on the Amazon Beauty data,  our method utilizes fewer {\em also-bought}, {\em produced-by} and {\em $<$null$>$} relations, while it uses other relations more frequently, especially the {\em bought-together} relation. We believe one reason for the success of our method is the diverse use of different relation types for discovering new path types.


\section{Conclusion}
\label{sec:conclusion}
We proposed a framework for improving the performance of path-based reasoning using reinforcement learning. Our results show that incorporating the heterogeneous context and the local neighborhood information results in a better performance for the query answering task. Our analysis shows that the type information is important for faster convergence and finding more diverse paths, and the neighborhood information improves the performance on unseen queries. In the future, we plan to explore more efficient strategies for action-space pruning to improve the scalability of existing RL solutions. Furthermore, we plan to develop more effective type embeddings considering the hierarchical structure of the type information.

\bibliographystyle{splncs04}
\bibliography{refs}

\end{document}